\documentclass[10pt,twocolumn,letterpaper]{article}

\usepackage[pagenumbers]{cvpr} 

\usepackage{graphics} 
\usepackage{epsfig} 
\usepackage{mathptmx} 
\usepackage{times} 
\usepackage{amsmath} 
\usepackage{amssymb}  
\usepackage{pifont}

\usepackage{xurl}
\usepackage{algorithm}
\usepackage{algpseudocode}
\usepackage{multirow}
\usepackage{booktabs}
\usepackage{caption}
\usepackage{color}

\newcommand{\ours}{SynDiff-AD}

\def\NoNumber#1{{\def\alglinenumber##1{}\State #1}\addtocounter{ALG@line}{-1}}

\newcommand{\model}{f}
\newcommand{\inputdata}{x}
\newcommand{\modelparameters}{\theta}
\newcommand{\outputdata}{y}
\newcommand{\loss}{M}
\newcommand{\datadistribution}{P}
\newcommand{\alldomain}{\mathcal{Z}}

\newcommand{\domain}{z}

\newcommand{\domainprobability}{g}
\newcommand{\dataset}{\mathcal{D}}
\newcommand{\captionsym}{c}
\newcommand{\prompt}{p}
\newcommand{\synt}{\mathrm{synth}}
\newcommand{\controlnet}{\theta_{\mathrm{CNet}}}
\newcommand{\vlm}{\theta_{\mathrm{vlm}}}
\newcommand{\cliptext}{\theta_{\mathrm{CLIP}}^{\mathrm{txt}}}
\newcommand{\clipimage}{\theta_{\mathrm{CLIP}}^{\mathrm{img}}}

\newcommand{\aug}{\mathrm{aug}}
\newcommand{\augmenteddataset}{\dataset_{\aug}}
\newcommand{\syntheticdataset}{\dataset_{\synt}}
\newcommand{\syntheticinputdata}{\inputdata_{\synt}}
\newcommand{\cmark}{\ding{51}}
\newcommand{\xmark}{\ding{55}}
\newcommand{\prediction}{\hat{y}}

\definecolor{cvprblue}{rgb}{0.21,0.49,0.74}
\usepackage[pagebackref,breaklinks,colorlinks,allcolors=cvprblue]{hyperref}

\def\paperID{9978} 
\def\confName{CVPR}
\def\confYear{2025}

\title{\ours: Improving Semantic Segmentation and End-to-End Autonomous Driving with Synthetic Data from Latent Diffusion Models}

\author{
Harsh Goel$^{1}$ \quad Sai Shankar Narasimhan$^{1}$ \quad Oguzhan Akcin$^{1}$ \quad Sandeep Chinchali$^{1}$ \\
$^1$The University of Texas at Austin\\
}

\begin{document}

\twocolumn[{%

\renewcommand\twocolumn[1][]{#1}%
\maketitle
\vspace{-3em}
\begin{center}
    \centering
    \captionsetup{type=figure}
    \includegraphics[width=1.0\textwidth]{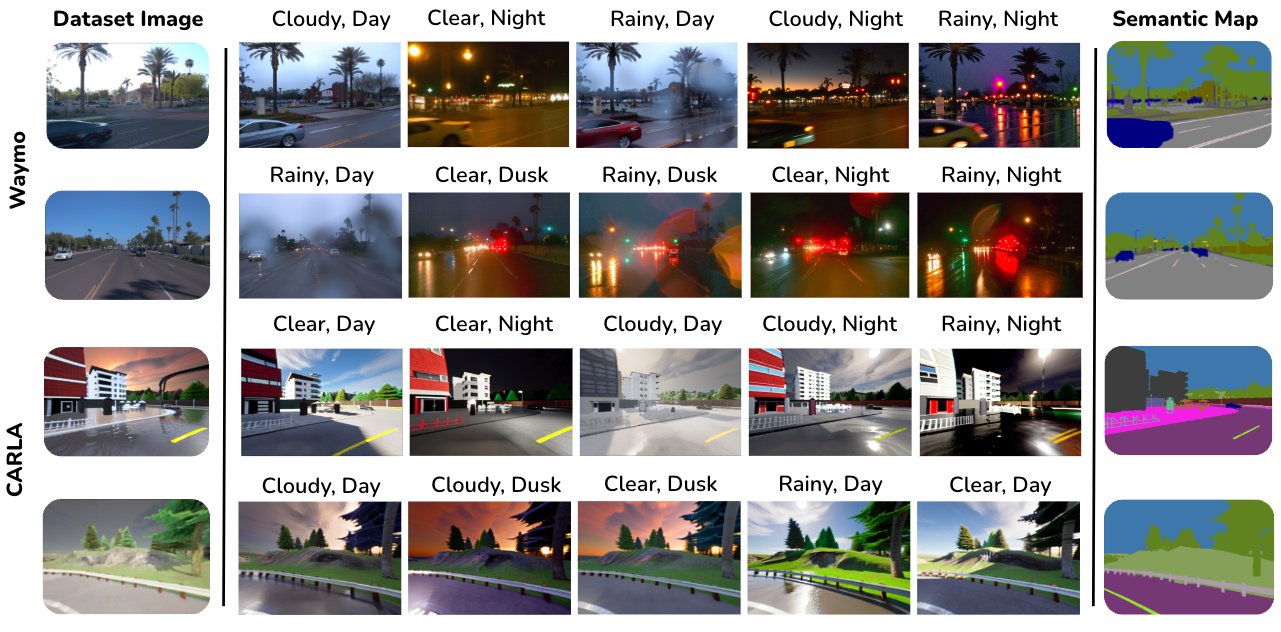}
    \caption{\textbf{Qualitative visualizations of synthetic images generated by \ours\ across various subgroups for the semantic segmentation (Waymo) and autonomous driving tasks (CARLA)}. We qualitatively visualize synthetic variants for an image in the original dataset (first column) and its corresponding semantic mask (last column). The original dataset images belong to over-represented data subgroups, such as ``Cloudy, Day" and ``Clear, Day", for both the semantic segmentation and autonomous driving tasks. We leverage our data augmentation pipeline, \textbf{\ours}, to generate images for different under-represented subgroups (illustrated in columns 2 to 6).}
    \label{fig:teaser}
\end{center}
}]

\maketitle

\begin{abstract}
    In recent years, significant progress has been made in collecting large-scale datasets to improve segmentation and autonomous driving models. These large-scale datasets are often dominated by common environmental conditions such as ``Clear and Day'' weather, leading to decreased performance in under-represented conditions like ``Rainy and Night''. To address this issue, we introduce \ours, a novel data augmentation pipeline that leverages diffusion models (DMs) to generate realistic images for such subgroups. \ours \space uses ControlNet---a DM that guides data generation conditioned on semantic maps---along with a novel prompting scheme that generates subgroup-specific, semantically dense prompts. By augmenting datasets with \ours, we improve the performance of segmentation models like Mask2Former and SegFormer by up to 1.2\% and 2.3\% on the Waymo dataset, and up to 1.4\% and 0.7\% on the DeepDrive dataset, respectively. Additionally, we demonstrate that our \ours \space pipeline enhances the driving performance of end-to-end autonomous driving models, like AIM-2D and AIM-BEV, by up to 20\% across diverse environmental conditions in the CARLA autonomous driving simulator, providing a more robust model. We release our code and pipeline at \url{https://github.com/UTAustin-SwarmLab/SynDiff-AD}.

\end{abstract}

\section{Introduction}

Autonomous Driving (AD) has seen significant progress over the last decade, fueled by terabytes of driving data collected daily from fleets of vehicles~\cite{liu2024surveyautonomousdrivingdatasets}. This data trains deep neural networks (DNNs) for key tasks like object detection, semantic segmentation, and end-to-end (E2E) AD. These advances have been critical in pushing AD systems closer to real-world deployment. 

\begin{figure*}[t]
    \centering   
    \includegraphics[width=0.9\linewidth]{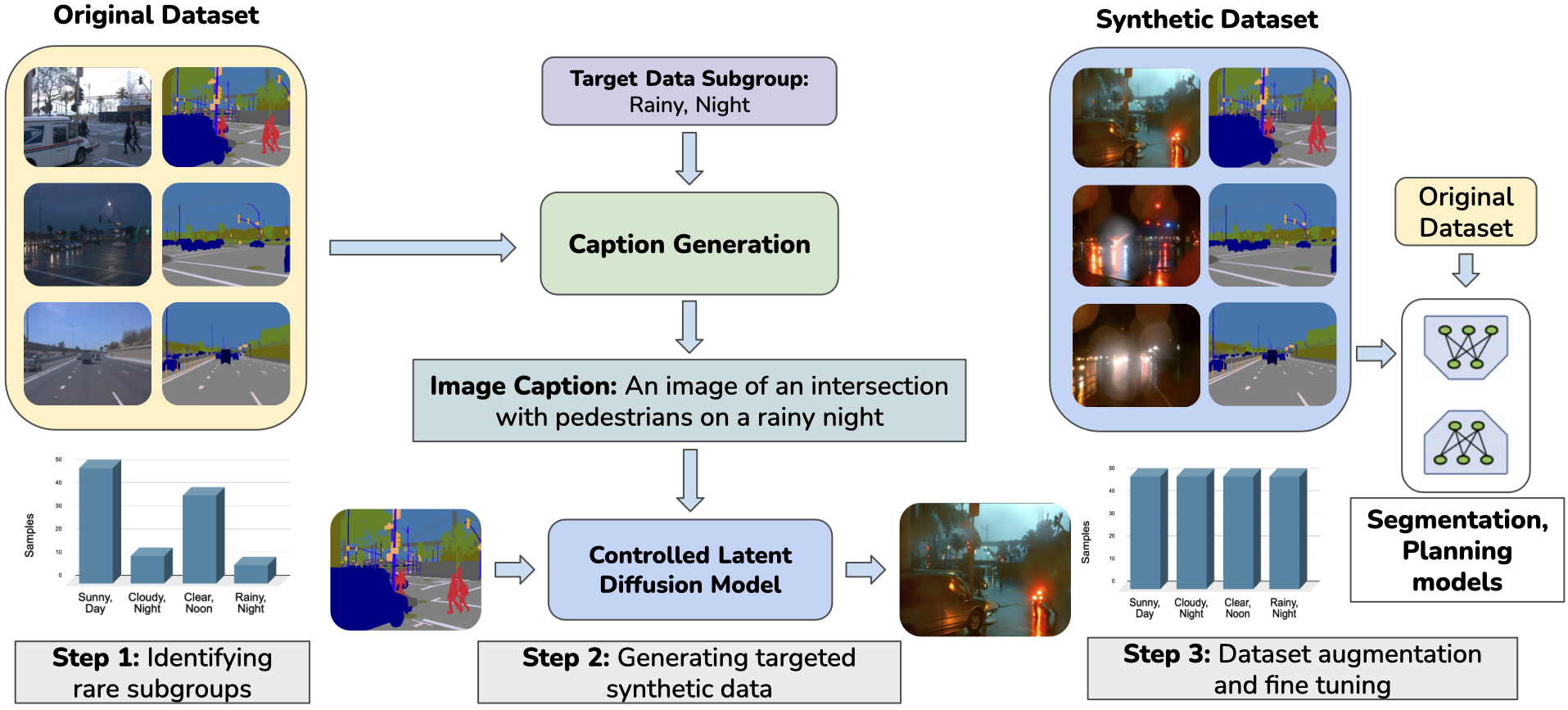}
    \captionsetup{font=small}
    \caption{\textbf{Our proposed synthetic data augmentation method -- \ours.
    } We begin by identifying subgroups of images in the dataset using a pre-trained vision-language model like CLIP~\cite{radford2021learning} (Step 1). Then, we create captions with our proposed caption generation scheme to synthesize images using a controlled latent diffusion model for under-represented subgroups   (Step 2). Finally, the generated synthetic data is used to augment the original dataset to fine-tune task-specific models (Step 3).}
    \label{fig:intro_fig}
    \vspace{-1.5em}
\end{figure*}

However, a major challenge persists: most of the data collected is biased toward common environmental conditions, such as sunny and clear weather. For example, the Waymo~\cite{sun2020scalability} and the DeepDrive~\cite{yu2020bdd100k} datasets comprise 65\% and 70\% of the samples collected during the day with sunny weather, while only $<$ 0.1\% and $<$ 0.01\% of the samples are collected during the night with rainy conditions, respectively. This imbalance leads to models that excel in common scenarios but underperform in rare scenarios. For instance, the performance of Mask2Former, a state-of-the-art (SOTA) semantic segmentation model, is 40\% lower on samples collected during night and rainy conditions when compared to those collected during the day and sunny weather conditions. Similar trends are observed in E2E AD driving models.


One natural solution commonly utilized in industry is to collect additional data for these under-represented conditions~\cite{akcin2023decentralized, akcin2023fal}. However, this is both costly and labor-intensive as the manual annotation of large datasets, particularly for complex tasks like semantic segmentation and E2E AD, significantly increases the cost and time required to improve model robustness. This necessitates a more scalable and efficient solution to address this challenge in AD datasets. 


We introduce \ours, a novel data augmentation pipeline that enhances the diversity of under-represented subgroups in AD datasets for semantic segmentation and E2E AD. 
Our method (Fig.~\ref{fig:intro_fig}) takes existing data samples and transforms their environmental conditions---such as converting sunny scenes to rainy ones---to boost the availability of data corresponding to rare scenarios. This transformation ensures the style of the image remains semantically consistent and preserves key annotations like segmentation targets. By leveraging latent diffusion models (LDMs)~\cite{rombach2021highresolution} to generate realistic, semantically consistent, and condition-altered data, \ours \space eliminates the need for costly manual labeling, enabling AD models to be fine-tuned on more balanced and comprehensive datasets.

Additionally, our approach introduces a novel prompting scheme inspired by textual inversion~\cite{gal2022image}. First, we generate detailed descriptions of over-represented data samples by leveraging LLaVA~\cite{zhang2023adding}, a vision-language captioning model. Then, by surgically altering these descriptions, we can guide the generation of synthetic data towards rare conditions. This enables us to balance AD datasets across all conditions. To this end, our key contributions are:
\begin{enumerate}
    \item We propose \ours, a conditional LDM-based approach that generates samples for under-represented subgroups in a dataset. \ours \space transforms samples from the over-represented subgroups without disturbing their semantic structure.
    \item Inspired by textual inversion, we introduce a novel prompting scheme to effectively steer the generation process in \ours \space towards under-represented subgroups.
    \item We empirically demonstrate that fine-tuning segmentation and AD models on real-world AD datasets---such as Waymo and DeepDrive---augmented with synthetic data from \ours, can improve their performance by up to 20\%, especially in under-represented data subgroups.
\end{enumerate}




\section{Background and Related Work}

\textbf{Semantic Segmentation.} Architecture modifications have been the primary focus of prior research in the semantic segmentation community. Initial studies utilized convolutional neural networks \cite{long2015fully, he2017mask} for segmentation. Currently, state-of-the-art semantic segmentation models \cite{cheng2022masked, liu2021swin,xie2021segformer,cheng2021per} utilize the transformer architecture \cite{vaswani2017attention}. In contrast to work centered on architectural improvements, this paper attempts to improve the performance of SOTA semantic segmentation models for autonomous driving with synthetic data augmentation via LDMs.

\textbf{End-to-End Autonomous Driving (E2E AD).}  E2E AD involves generating motion plans for autonomous vehicles conditioned on multiple sensor modalities such as RGB images from cameras, depth measurements from LIDAR~\cite{chitta2021neat,chitta2022transfuser, sobh2018end, natan2022end, chen2022learning}, etc. The primary focus of these methods has been on the design of model architectures to predict future waypoints. These methods depend on simulators to generate extensive training data for effective generalization across various weather conditions. In contrast, our proposed approach seeks to augment the training data with synthetic data for scenarios with limited data availability.

\textbf{Generative Models.} Text-to-image generative models, especially those using LDMs \cite{sohl2015deep} like Stable Diffusion \cite{rombach2021highresolution} and Glide \cite{nichol2022glide}, have opened up new possibilities for generating high-quality synthetic data conditioned on text. These models are trained on large datasets such as LAION~\cite{schuhmann2022laion5b}, which contain over 5 billion pairs of images and text.

However, it is difficult to generate images through such models by only modifying the appearance of certain prescribed subjects in a reference image. To address this issue, several methods~\cite{gal2022image,roich2022pivotal,ruiz2023dreambooth,zhang2023adding} have been introduced to adjust specific parts of the image. PTI~\cite{roich2022pivotal} introduced pivotal tuning for diffusion models to preserve in domain latent qualities, such as the facial structures of a human while changing its identity and appearance (e.g., skin tones). 

Textual Inversion~\cite{gal2022image, ruiz2023dreambooth} is another method to synthesize images that fit novel ideas, concepts, or subgroups through a small set of images provided by a user.
More recently, ControlNet~\cite{zhang2023adding} introduces further control of the text-to-image generation process using additional inputs such as semantic maps and depth maps. To better control image synthesis, FreestyleNet~\cite{xue2023freestyle} proposes a cross-rectified attention mapping between semantic maps and text prompts. Lastly, Edit-Anything~\cite{gao2023editanything} combines ControlNet with semantic maps obtained from the Segment Anything Model (SAM) \cite{kirillov2023segany} and an image-language grounded model such as BLIP2~\cite{li2023blip2} to generate high-quality images for different styles.

\textbf{Synthetic Data for Perception and Planning Tasks.} 
Prior works have used generative adversarial networks (GANs) to generate synthetic data to improve the generalization of perception models \cite{tanaka2019data, zhang2021datasetgan}. Currently, diffusion models are used to generate synthetic data for object classification \cite{he2022synthetic,azizi2023synthetic}, object detection \cite{lin2023explore}, and semantic segmentation \cite{zhang2023emit,yang2023freemask, trabucco2023effective, lei2023image, wu2024datasetdm, jia2023dginstyle, goel2024improving}. More recently, DatasetDM~\cite{wu2024datasetdm} proposes a method that learns to produce both perception annotations, such as segmentation maps, and synthetic images to improve segmentation models. Aligned with our work, DGInStyle~\cite{jia2023dginstyle} presents a method that learns to generate synthetic images across multiple domains for semantic segmentation. However, these methods don't address the quality of learned segmentation models on under-represented subgroups in the training data, where models perform disproportionately worse. 


For AD, generating synthetic data typically relies on high-fidelity 3D simulation engines like Unity and Unreal Engine (UE4). These simulators allow for the procedural generation of diverse conditions, such as weather and lighting, within controlled environments \cite{song2023synthetic}. However, such simulation tools come with significant computational and development costs, and they often require additional expert driving plans to guide the models. In contrast, our approach offers an efficient alternative by generating semantically consistent synthetic camera images for AD without relying on expensive simulators or human-crafted driving plans. Our method ensures that generated data remains aligned with the original annotations to balance the data distribution across all subgroups during training.

\section{Method}
\label{ss:method_assumption}
In this section, we describe our proposed approach, \ours, for synthetic data generation specific to under-represented subgroups.  


\underline{\emph{Notations:}} First, we will briefly describe the notations used in this section. We investigate a practical AD scenario where a vision or planning model is trained on a dataset dominated by common environmental conditions. This leads to relatively poorer performance in under-represented data subgroups, like rainy weather. In AD scenarios, a segmentation model or an E2E AD model $\model$ maps an image $\inputdata$ to an output $\prediction = \model(\inputdata ; \modelparameters)$, where $\modelparameters$ represents the model parameters. Note that $\prediction$ is either a segmentation map or a pose offset (for E2E AD). The model $\model$ is trained on a dataset $\dataset = \{(\inputdata_i, \outputdata_i)\}_{i=1}^N$ consisting of $N$ samples. The pair ($\inputdata_i$, $\outputdata_i$) belongs to an unknown subgroup $\domain$. The set of all data subgroups, $\alldomain$, is finite, and each element $\domain \in \alldomain$ is described in natural language similar to \cite{metzen2023identification}. Therefore, the data distribution $\datadistribution$, from which the dataset $\dataset$ is drawn, is a mixture of subgroup-specific data distributions $\datadistribution_{\domain}$ for all $\domain \in \alldomain$. The overall data distribution is given by $\datadistribution = \sum_{\domain \in \alldomain} \domainprobability_\domain \datadistribution_{\domain}$, where $\domainprobability_\domain$ is the mixture probability of $\domain$ in the data distribution $\datadistribution$. The notation $U({z_1, \dots, z_n})$ for any $n$ indicates the uniform probability distribution over the set of $n$ elements ${z_1, \dots, z_n}$.

The performance of the model $\model$ is measured by a metric $\loss(\model(\inputdata; \modelparameters), \outputdata)$, which can be the mean Intersection over Union (mIoU) \cite{chitta2022transfuser} for segmentation or the Driving Score (DS) for E2E AD \cite{chitta2021neat}. \textbf{Our goal} is to maximize the performance for all subgroups $\domain \in \alldomain$.

Our key intuition is to synthesize realistic data for under-represented subgroups that semantically resemble data from over-represented subgroups. This leads to the availability of high-quality labeled data (as we can use the corresponding labels from over-represented subgroups) for under-represented subgroups. To achieve this, we rely on the recent advancements in the controlled image generation paradigm, such as ControlNet~\cite{zhang2023adding}. 

In our approach, we first fine-tune ControlNet with subgroup-specific semantically dense captions (Sec.~\ref{ss:image_gen}). Our approach to obtaining such dense captions is explained in Sec.~\ref{ss:prompting}. Then, to specifically target data generation for under-represented subgroups, we surgically alter these captions obtained from images corresponding to over-represented subgroups (Sec.~\ref{ss:gen}). Both controlled image generation using ControlNet \cite{zhang2023adding} and the proposed caption generation scheme ensure that the semantic structure of the generated image is preserved, thereby eliminating the need for human annotation.

\subsection{Controlled Image Generation}
\label{ss:image_gen}
\ours \space leverages semantic mask labels and text-to-image conditional generative models to generate pixel-aligned image and semantic mask pairs for under-represented subgroups. This process utilizes ControlNet~\cite{zhang2023adding}, which is built on top of Stable Diffusion~\cite{rombach2021highresolution}. ControlNet incorporates a separately trained U-Net~\cite{olaf2015unet} encoder that takes semantic masks as additional input, in addition to the LDM. We use the notation $\controlnet(x|y,c)$ for ControlNet in this paper. Here, $x$ refers to the generated image, $y$ refers to the semantic mask input to the U-Net encoder, and $c$ refers to the caption input or text prompt to the LDM. Though ControlNet can generate images that align with semantic mask inputs, it is not trained specifically for scenario generation pertaining to AD. Therefore, we fine-tune the additional U-Net parameters of ControlNet with semantic masks from AD datasets. Note that the base LDM's model parameters remain fixed. Additionally, \ours \space supervises ControlNet with text-based image captions enhanced by integrating the specific data subgroup of the image to be reconstructed, as a stylistic element. We describe the process to obtain these captions in the next section (Sec.~\ref{ss:prompting}).

Finally, during generation, ControlNet guides the denoising process to synthesize images that align with the semantic layout of the input masks. Additionally, through semantically dense captions that include under-represented subgroups as described in Sec.~\ref{ss:prompting}, we enable the generation of synthetic data tailored for such subgroups for the subsequent fine-tuning of the task-specific models. In the next section, we describe how we obtain targeted image captions that are integrated with the required subgroup information for high-quality subgroup-specific image generation.

\subsection{Subgroup-Specific Caption Generation (CaG)}
\label{ss:prompting}
\begin{figure}[t]
    \centering   
    \includegraphics[width=\linewidth]{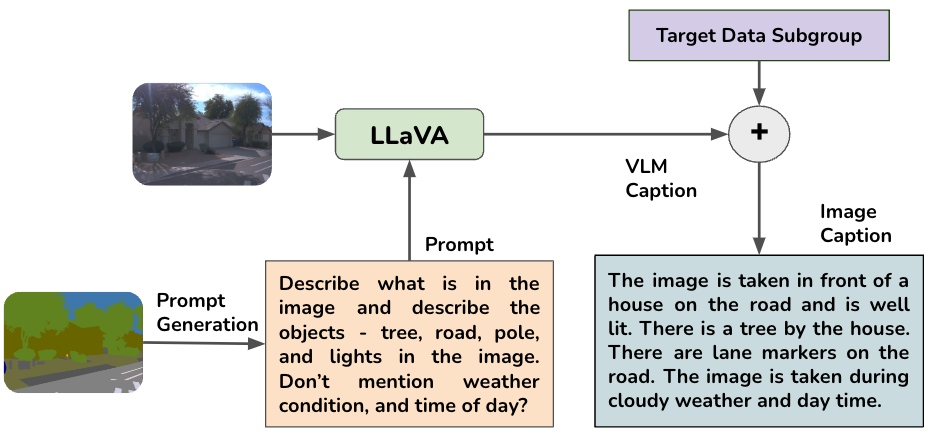}
    \captionsetup{font=small}
    \caption{\textbf{Caption Generation Pipeline.} Our proposed caption generation scheme is used to fine-tune ControlNet and subsequently generate synthetic data. We query LLaVA with a prompt constructed from the semantic mask to obtain a caption for its corresponding image. A subgroup, either associated with the image or a target under-represented condition, is appended to this caption for fine-tuning ControlNet or image synthesis, respectively.}
    \label{fig:cap_gen_pipeline}
    \vspace{-1em}
\end{figure}
We enhance image generation using ControlNet through subgroup-specific semantically dense captions. As mentioned in Sec~\ref{ss:image_gen}, we also use these targeted captions during the fine-tuning process of ControlNet. Our key intuition is that such targeted captions are concatenations of semantic information and subgroup-specific information, and these captions aid in retaining the correct semantic structure so that the generated image effectively aligns with the segmentation mask input. 

Our approach uses large vision-language models (VLMs) to precisely capture semantic information. These have been used in prior works~\cite{gao2023editanything, azizi2023synthetic} to enhance the realism of generated images. \ours \space employs LLaVA~\cite{liu2023visual}, a VLM that first extracts prompt-specific image features using CLIP from an image input. These features are projected into the token space of a Large Language Model (LLM) (in this case, Vicuna \cite{vicuna2023}) to produce captions. In this work, we represent a VLM as $\vlm(x,p)$, where $x$ is the input image, and $p$ is the input prompt. The prompt $p$ is constructed based on the object classes extracted from parsing the semantic masks. Additionally, we instruct LLaVA not to include keywords related to the given subgroup $\domain$. For example, a sample query would be: ``Provide a description of the objects - [\textit{from semantic masks}] - and their relationships with respect to each other. Describe the background of the scene and image quality. Do not mention the subgroups - [$\domain  \in \alldomain$]''. This is depicted in Fig.~\ref{fig:cap_gen_pipeline}.

Next, we augment the caption generated by the VLM with subgroup-specific information. For example, if the subgroup is ``Cloudy, Night'', we add the line ``Image taken in cloudy weather at night time'' to the caption from LLaVA.  Generally, the subgroup information is not available apriori. To identify the data subgroup of an image, we rely on pre-trained CLIP models. We represent the CLIP text and image encoders as $\cliptext$ and $\clipimage$, respectively. As stated earlier, we assume that the set of all subgroups $\alldomain$ is described in natural language. For a specific image $\inputdata$ in the dataset, we first obtain the CLIP image embeddings $\clipimage(x)$ for the image and the CLIP text embeddings $\cliptext(u)$ for all the subgroups in $u \in \alldomain$. Finally, we assign the subgroup $\domain$ to the image $\inputdata$ as $z = \mathrm{argmax}_u \; \cliptext(u)^T\clipimage(x)$. 

Once we obtain the target prompt by concatenating the caption obtained from LLaVA with subgroup-specific information, we use it to fine-tune ControlNet and subsequently synthesize subgroup-specific images.

\subsection{Generating Synthetic Dataset}
\label{ss:gen}
\textbf{Augmenting Segmentation Datasets with \ours.}
We now describe the augmentation process utilized in \ours \space in Algorithm~\ref{alg:ss_data_augmentation} to obtain $N_{synth}$ synthetic samples. Initially, we randomly pick an image $\inputdata$ and the corresponding semantic map $\outputdata$ from $\dataset$ (line 6). Following this, we obtain the subgroup $\domain$ for $x$ using CLIP, as explained in Sec.~\ref{ss:prompting} (line 7). Subsequently, we generate a caption $\captionsym$ for the image $\inputdata$ using LLaVA with a prompt $\prompt$ derived from the semantic mask $\outputdata$ and subgroup $\domain$ (lines 8 and 9). Next, we uniformly sample a target subgroup $\domain^* \in \alldomain\ \setminus \{\domain\}$ for data synthesis (indicated by $U(\alldomain \setminus \{\domain\})$ in line 10). Here, we ensure that we do not synthesize images for the same subgroup $z$ corresponding to the image $x$. We generate a caption $\captionsym^*$ by combining the VLM-generated caption $\captionsym$ with the target subgroup $\domain^*$, as described in Section~\ref{ss:prompting} (line 11). This combined caption serves as the prompt for synthesizing a new image $\syntheticinputdata$ from ControlNet -- $\controlnet(\inputdata|\outputdata,\captionsym^*)$ (line 12). This enables us to construct $\syntheticdataset$ that comprises image-segmentation annotation pairs $(\syntheticinputdata, \outputdata)$ to create the augmented dataset $\augmenteddataset$ for further fine-tuning (line 15).
\begin{algorithm}
\caption{Synthetic Data Augmentation Algorithm}\label{alg:ss_data_augmentation}
\begin{algorithmic}[1]
\State \textbf{Input:} Dataset $\dataset = \{(\inputdata_i,\outputdata_i)\}^N_{i=1}$ with images $\{\inputdata_i\}_{i=1}^N$, and corresponding semantic masks $\{\outputdata_i\}_{i=1}^N$, set of all subgroups $\alldomain$, CLIP text encoder $\cliptext$, CLIP image encoder $\clipimage$, ControlNet $\controlnet$, VLM $\vlm$, number of synthetic images to be generated $N_{\synt}$. 
\State \textbf{Output:} Augmented dataset $\augmenteddataset$
\Procedure{Data Augmentation}{$\dataset$}
    \State $\syntheticdataset \gets \emptyset$
    \For{$i = 1$ to $N_{\text{synth}}$}
        \State Sample the pair $(\inputdata, \outputdata) \in \dataset$
        \State Obtain $\domain \gets \mathrm{argmax}_{u \in \alldomain}(\cliptext(u)^T\clipimage(x))$    \NoNumber{\Comment{Subgroup Identification via CLIP}}
        \State Obtain prompt $\prompt$ based on $\outputdata$ and $\domain$
        \State $\captionsym \gets \vlm(\inputdata,\prompt)$     \; 
        \NoNumber{\Comment{Generate caption $\captionsym$ using the VLM}}
        \State Select target subgroup $\domain^* \sim U(\alldomain \setminus \{\domain\})$ \;\;  
        \NoNumber{\Comment{Sample target subgroup uniformly}}
        \State $\captionsym^* \gets \captionsym \bigoplus \domain^* $ \;  \Comment{Add target subgroup as style}
        \State Generate synthetic image $\syntheticinputdata \gets  \controlnet(\inputdata | \outputdata, \captionsym^*)$  
        \State $\syntheticdataset \gets \syntheticdataset \cup \{(\syntheticinputdata, \outputdata)\}$ 
    \EndFor
    \State $\augmenteddataset \gets \dataset \cup \syntheticdataset$
    \State \textbf{return} $ \augmenteddataset$
\EndProcedure
\end{algorithmic}

\end{algorithm}

\begin{table}[ht]
    \centering
    \begin{subtable}[t]{\textwidth}
        \begin{tabular}{lccccc}
            \hline
            \multirow{2}{*}{\textbf{Augmentation}} & 
            \multicolumn{2}{c}{\textbf{Waymo}}& \multicolumn{2}{c}{\textbf{DeepDrive}}\\
            \cmidrule(lr){2-3} \cmidrule(lr){4-5}
            & \textbf{mIoU} &\textbf{mF1} & \textbf{mIoU} &\textbf{mF1} \\
            \hline
             -                & 47.5  & 60.0  & 54.6  & 70.5  \\
             Color Aug             & 47.8  & 62.4  & 54.6  & 70.6  \\
             Ours             & 47.7  & 62.4  & \textbf{55.5}  & 71.7 \\
             Color Aug + Ours      & \textbf{48.3}  & \textbf{63.1}  & \textbf{55.5}  & \textbf{71.8}  \\
            \hline
        \end{tabular}
        \captionsetup{
      justification=raggedright, font=small, singlelinecheck=false}

        \caption{Results for Mask2Former with Resnet-50 backbone.}
    \end{subtable}
    
    \vspace{0.5cm} 
    
    \begin{subtable}[t]{\textwidth}

        \begin{tabular}{lccccc}
            \hline
            \multirow{2}{*}{\textbf{Augmentation}} & 
            \multicolumn{2}{c}{\textbf{Waymo}}& \multicolumn{2}{c}{\textbf{DeepDrive}}\\
            \cmidrule(lr){2-3} \cmidrule(lr){4-5}
            & \textbf{mIoU} &\textbf{mF1} & \textbf{mIoU} &\textbf{mF1} \\
            \hline
             -                & 48.0  & 59.5  & 57.3  & 73.2  \\
             Color Aug             & 48.2  & 57.8  & 57.6  & 73.7  \\
             Ours             & \textbf{49.2}  & 62.0  & 56.9  & 73.0  \\
             Color Aug + Ours      & \textbf{49.2} & \textbf{62.2}  & \textbf{58.7}  & \textbf{74.8}  \\
            \hline
        \end{tabular}
            \captionsetup{
      justification=raggedright, font=small, singlelinecheck=false}
        \caption{Results for Mask2Former with Swin Transformer backbone.}
    \end{subtable}
    
    \vspace{0.5cm} 
    
        \begin{subtable}[t]{\textwidth}

        \begin{tabular}{lccccc}
            \hline
            \multirow{2}{*}{\textbf{Augmentation}} & 
            \multicolumn{2}{c}{\textbf{Waymo}}& \multicolumn{2}{c}{\textbf{DeepDrive}}\\
            \cmidrule(lr){2-3} \cmidrule(lr){4-5}
            & \textbf{mIoU} &\textbf{mF1} & \textbf{mIoU} &\textbf{mF1} \\
            \hline
             -                & 45.3  & 61.2  & 54.7  & 70.7  \\
             Color Aug             & 45.5  & 61.4  & 55.0  & 71.1  \\
             Ours             & 46.9  & 63.7  & \textbf{55.4}  & \textbf{71.6}  \\
             Color Aug + Ours      & \textbf{47.6}  & \textbf{64.3}  & 54.7  & 70.9  \\
            \hline
        \end{tabular}
        \captionsetup{
      justification=raggedright, font=small, singlelinecheck=false}
        \caption{Results for Segformer with the MiT-B3 backbone.}
    \end{subtable}
    \captionsetup{font=small}
    \caption{\textbf{Segmentation models trained on the augmented dataset with \ours\ outperform those trained on original data}. We compare the performance of models like Mask2Former and Segformer across various training setups: no augmentation, color augmentation, augmentation with \ours, and color augmentation combined with \ours. Our results demonstrate that models trained on color-augmented datasets, supplemented with synthetic data generated by \ours, consistently outperform all other variants. Finally, we notice that only using \ours\ for augmentation also improves the performance of the models.}
    \label{tab:performance_per_npm}
    \vspace{-1em}

\end{table}

\textbf{Synthetic Dataset for Autonomous Driving with \ours.}
The procedure for generating synthetic images for autonomous driving follows a similar approach as proposed in Algorithm~\ref{alg:ss_data_augmentation}, with some specific modifications. First, the annotations $\outputdata$ for AD tasks represent future offsets or waypoints. Therefore, to ensure compatibility for fine-tuning AD models, we generate synthetic data that preserves the semantic layout of an observed scene. This can be done by utilizing the available annotated semantic masks. We can readily obtain semantic maps from the simulator since we use the CARLA~\cite{Dosovitskiy17} simulation environment. Alternatively, we can also utilize semantic segmentation models to derive masks from the images in real-world AD data for the subsequent data synthesis.

\section{Experiments and Discussions}
\label{ss:experiments}


In this section, we evaluate our \ours \space pipeline on two key tasks: semantic segmentation and AD, focusing on the impact of synthetic data on the model's performance. We first analyze the overall performance of segmentation and AD models, followed by an investigation of how synthetic data influences subgroup-specific performance. Lastly, we assess the effectiveness of our caption generation scheme on the quality of synthetic data.

\textbf{Models.}
We have used two state-of-the-art semantic segmentation models: Mask2Former~\cite{cheng2022masked} with the Swin Transformer (Swin-T)~\cite{liu2021swin} and ResNet-50~\cite{he2016deep} backbones, and SegFormer with MIT-B3~\cite{xie2021segformer} backbone. For AD tasks, we have used AIM-2D and AIM-BEV \cite{chitta2021neat} models. Both these models are trained on a single-camera view to predict future waypoints on both the original and augmented datasets.

\textbf{Datasets.}
We have used the Waymo \cite{sun2020scalability} and the DeepDrive \cite{yu2020bdd100k} datasets for the semantic segmentation task. These datasets are categorized into subgroups based on weather (\textit{Rain, Clear, Cloudy}) and time of day (\textit{Dawn / Dusk, Morning, Night}), resulting in a total of nine subgroups. For segmentation, the training data is heavily biased towards ``Clear, Day'' and ``Cloudy, Day'' subgroups, which make up more than 80\% of the data.

For AD experiments, we have used a dataset collected from CARLA~\cite{Dosovitskiy17} using the expert driving policy from NEAT~\cite{chitta2021neat}. The dataset includes images from three towns and 15 driving routes under varied environmental conditions. Notably, a key deviation in our dataset collection compared to NEAT is that we fix weather conditions for each route to replicate any real AD dataset collection. Like the segmentation experiments, the dataset is categorized into subgroups based on weather and time of day. Finally, the trained models are tested in 3 towns over 27 routes. 
\begin{table}
    \centering
    \captionsetup{ singlelinecheck=false}
    \begin{tabular}{@{}cccccc@{}}
        \toprule
        \textbf{Model} & \textbf{Data}  & \textbf{RC $\uparrow$} & \textbf{IS $\uparrow$}& \textbf{DS $\uparrow$} \\
        \midrule
         AIM-2D \cite{chitta2022transfuser}& Real     & 64.61 & 0.388 & 21.64 \\
         AIM-2D \cite{chitta2022transfuser} & Real + Syn    & \textbf{71.86} & \textbf{0.458} & \textbf{34.36}  \\
         \noalign{\smallskip}
         \cline{2-5}
         \noalign{\smallskip}
         AIM-BEV \cite{chitta2022transfuser}& Real     & 71.35 & 0.419 & 28.61 \\
         AIM-BEV \cite{chitta2022transfuser}& Real + Syn      & \textbf{73.57} & \textbf{0.478} & \textbf{35.86}  \\
         \noalign{\smallskip}
         \cline{2-5}
         \noalign{\smallskip}
         Expert & - & 77.4 & 0.772 & 54.6\\
         \bottomrule
    \end{tabular}
    \captionsetup{font=small}
         \caption{\textbf{Autonomous driving performance improves with augmented training data}. We show that AD models fine-tuned on augmented data have improved Route Completion, Infraction, and Driving Scores. The values are averaged over 3 runs.}
    \label{tab:performance_per_npmad}
    \vspace{-1em}
\end{table}

\begin{table*}[htbp]
    \centering
    \captionsetup{ singlelinecheck=false}
\label{tab:transposed} 
\begin{tabular}{@{}lcccccccccc@{}}
\toprule
\multirow{2}{*}{\textbf{Model}} & \multirow{2}{*}{\textbf{Aug}}& 
\multicolumn{3}{c}{\textbf{Clear}}& \multicolumn{3}{c}{\textbf{Cloudy}}&\multicolumn{3}{c}{\textbf{Rain}}\\
\cmidrule(lr){3-5} \cmidrule(lr){6-8} \cmidrule(lr){9-11}
& & \textbf{Twilight} &\textbf{Day} & \textbf{Night} &\textbf{Twilight} & \textbf{Day} & \textbf{Night} & \textbf{Twilight} & \textbf{Day} & \textbf{Night} \\
\midrule
\textbf{AIM-2D}& \xmark & 19.69 & 23.20 & 6.11 & \textbf{37.25} & 18.77 & 43.72 & 14.68 & 23.93 & 3.42\\ 
\textbf{AIM-2D} & \cmark & \textbf{39.04} & \textbf{40.30} & \textbf{29.02} & 19.11 & \textbf{33.44} & \textbf{46.32} & \textbf{16.68} & \textbf{50.94} & \textbf{34.23} \\ 
         \midrule
\textbf{AIM-BEV}& \xmark & 39.78 & 31.42 & 2.73 & \textbf{29.88} & \textbf{44.68} & \textbf{43.22} & 18.07 & 43.73 & 3.97\\ 
\textbf{AIM-BEV} & \cmark & \textbf{58.37} & \textbf{47.94} & \textbf{25.42} & 14.93  & 44.22 & 35.03 & \textbf{27.52} & \textbf{53.67} & \textbf{15.64} \\ 
\bottomrule
\end{tabular}
\captionsetup{font=small}
    \caption{\textbf{AD models trained on augmented datasets exhibit improved driving scores.} We show that AD models fine-tuned on augmented datasets (indicated by Aug) have higher driving scores, especially over rare subgroups (e.g., during ``Rain, Night'' and ``Rain, Day'') where the models trained on the original dataset underperform. Also, note that the best-performing AD model for each subgroup (except ``Cloudy, Twilight'' and ``Cloudy, Day'') is a model variant fine-tuned on augmented data.}
    \label{tab:performance_per_subgroup_sem_ad}
    \vspace{-1em}
\end{table*}

\begin{table}
    \centering
        
     \begin{small}
    \begin{subtable}[t]{0.6\textwidth}
            \captionsetup{
       font=small, singlelinecheck=false,labelformat=empty}

        \caption{\textbf{Waymo}}
        \begin{tabular}{@{}cc|ccc@{}}
            \toprule
            \textbf{Subgroup} & \textbf{-} & \textbf{CAug} & \textbf{Ours} & \textbf{CAug + Ours} \\
            \midrule
              Clear, Dawn / Dusk & 40.5 & 41.1 & \textbf{43.0} & 42.4  \\
              Clear, Day       & 49.8 & \textbf{49.9} & 49.6 & 49.4 \\
              Clear, Night     & 36.6 & 33.4 & \textbf{38.2} & 38.0 \\
              Cloudy, Dawn / Dusk& 45.4 & 47.2 & \textbf{51.6} & 49.4  \\
             Cloudy, Day &    55.5 & 55.4& \textbf{56.7} & 56.1 \\
              Cloudy, Night    & 38.2 & 36.0 & 38.2 & \textbf{42.3}  \\
             Rain, Dawn / Dusk   & 37.2 & 38.4 & \textbf{42.2} & \textbf{42.2}   \\
              Rain, Day        & 49.2 & 49.2 & \textbf{49.4} & 49.2 \\
              Rain, Night      & 34.4 & 33.3 & \textbf{35.1} & 34.8  \\
            \bottomrule
        \end{tabular}

    \end{subtable}
    \vspace{0.5cm}
    \begin{subtable}[t]{0.6\textwidth}
        \captionsetup{
     font=small, singlelinecheck=false,labelformat=empty}
    \caption{\textbf{DeepDrive}}
        \begin{tabular}{@{}cc|ccc@{}}
            \toprule
             \textbf{Subgroup} & \textbf{-} & \textbf{CAug} & \textbf{Ours} & \textbf{CAug + Ours} \\
               \midrule
              Clear, Dawn / Dusk & 47.8 & 47.5  &  46.6 & \textbf{51.7}\\
              Clear, Day         & 56.4 & 56.7  &  56.2 & \textbf{57.7} \\
              Clear, Night       & 53.4 & 53.8  &  50.8 & \textbf{64.1}\\
              Cloudy, Dawn / Dusk& \textbf{48.1} & 42.6  &  42.6 & 37.7 \\
              Cloudy, Day        & \textbf{60.7} & 60.4 &  51.3 & 59.3\\
              Cloudy, Night      & 55.7 & \textbf{56.6}  &  51.5 & 48.7\\
             Rain, Dawn / Dusk   & 49.9 & 49.6  &  44.3 & \textbf{54.5} \\
              Rain, Day          & 56.3 & 58.4  &  56.3 & \textbf{59.3}\\
              Rain, Night        & 31.4 & 33.2  &  31.0 & \textbf{48.7}\\
            \bottomrule
        \end{tabular}

    \end{subtable}
    \end{small}
    \vspace{-2em}
    \captionsetup{font=small}
    \caption{\textbf{Improvement in mIoU over different data subgroups with synthetic data augmentation by \ours}. We train Mask2Former Swin-T (the best-performing model in Table~\ref{tab:performance_per_npm}) on the original dataset and the augmented dataset with \ours, with and without color-augmentation (CAug).  We show that the Mask2Former model trained on the augmented Waymo dataset with \ours\ exhibits superior performance to the model trained only on the original dataset on most subgroups except ``Clear, Day''. We see similar trends in the DeepDrive dataset. }
    \label{tab:performance_per_subgroup_sem}
    \vspace{-2em}
\end{table}
\begin{figure*}[ht]
    \centering   
    \captionsetup{font=small}
    \includegraphics[width=\linewidth]{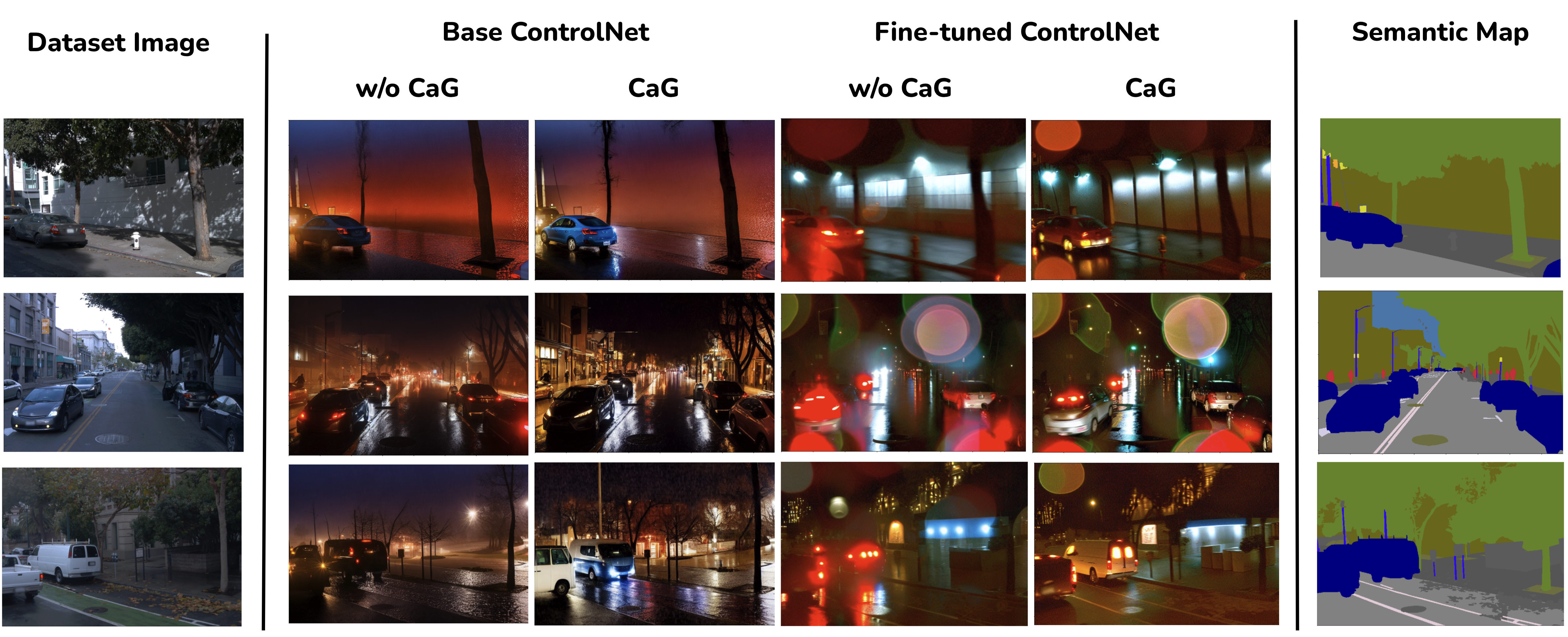}
    \vspace{-2em}
    \caption{\textbf{Ablation of synthesized images on the Waymo dataset}. We qualitatively visualize synthetic images for ``Rainy and Night'' weather for an image from the Waymo dataset taken in ``Clear and Day'' conditions. We generate images from a base ControlNet model and a fine-tuned ControlNet model, with text prompts obtained with and without our subgroup-specific caption generation scheme (CaG). First, as seen from images in columns 1 and 2, the base ControlNet model yields synthetic images that are hyper-realistic and not dataset-specific. The fine-tuned ControlNet model generates more dataset-specific images as observed by artifacts such as the blurring due to raindrops in rainy weather and realistic backgrounds. Additionally, we show that CaG further improves the quality of synthesized images. For instance, images generated without CaG using a fine-tuned ControlNet in rows 2 and 3 show unnatural generative artifacts on cars with missing pavements and lane markers. These artifacts are absent in the synthesized image via CaG. }
    \label{fig:syn_res}
    \vspace{-1em}
\end{figure*}
\textbf{Experiment Setup.}
For the segmentation task, we first fine-tune ControlNet for 10 epochs with a constant learning rate of $1 \times 10^{-5}$ with text-based prompts constructed via our caption generation method on both the Waymo and DeepDrive datasets. Then, we generate approximately $36,000$ synthetic samples for the Waymo dataset and $9,000$ samples for the DeepDrive dataset via \ours. These are distributed equally across all subgroups. Finally, the segmentation models are pre-trained for $80,000$ iterations on the original datasets and then fine-tuned for an additional $80,000$ iterations on the augmented datasets with batch size 4. This enables us to assess the effect of synthetic data on the performance of the model.

For AD, we fine-tune ControlNet to the dataset collected from CARLA for 20 epochs with a constant learning rate of $1 \times 10^{-5}$. During ControlNet inference, we swap the style as discussed in Algorithm~\ref{alg:ss_data_augmentation} to obtain the synthetic images for under-represented subgroups. The synthesis and fine-tuning are conducted over 8 A5000 GPUs. Finally, we train AIM-2D and AIM-BEV \cite{chitta2021neat}. These models, when fine-tuned respectively with both datasets, have an equivalent number of gradient updates during training. Hence, we increase the batch size when training on augmented datasets. These experiments are conducted over a single A5000 GPU.

\textbf{Evaluation Metrics.}
We evaluate the performance of the segmentation models using the mean Intersection over Union (mIoU) and the mean F1 (mF1) scores. Note that these metrics are averaged across all semantic classes in the datasets or subgroup-specific datasets. For AD models, we evaluate the models on the primary metric---Driving Score (DS)---which combines the Route Completion (RC) and Infraction Score (IS) \cite{chitta2021neat, chitta2022transfuser}. Route Completion (RC) refers to the fraction of the route completed by the vehicle, and the Infraction Score (IS) is inversely correlated with the violations made by the vehicle, such as collisions, lane infractions, red lights, and stop sign violations.
 
We now discuss the results obtained from our experiments by asking the following research questions (RQ).

\textbf{RQ1: Does augmenting with synthetic data using \ours\ improve the overall performance?}

We evaluate the performance of the segmentation models and AD models on the held-out test data. For semantic segmentation, we compare and contrast \ours\ based augmentation with color-based augmentation--which distorts the hue, saturation, and brightness of the image. The results in Tables~\ref{tab:performance_per_npm} show that fine-tuning on color-augmented datasets, supplemented with synthetic samples from \ours, results in the most performant models. For instance on the Waymo Dataset, compared to the model on trained the original dataset without and with color augmentation, we observe improvements in SegFormer (MiT-B3) by a substantial \textbf{+2.3 mIoU} and \textbf{+2.1 mIoU} respectively, Mask2Former (Resnet-50) by \textbf{+0.8 mIoU} and \textbf{+0.5 mIoU} respectively, and Mask2Former (Swin-T) by \textbf{+1.2 mIoU} and \textbf{+1.0 mIoU} respectively. Similar trends are seen on the DeepDrive dataset, with a notable \textbf{+1.4 mIoU} gain for Mask2Former (Swin-T). Additionally, as shown in Table ~\ref{tab:performance_per_npm}, we can also improve the performance of segmentation models by solely using \ours\ as the data augmentation method.

For AD experiments in Table ~\ref{tab:performance_per_npmad}, we observe improvements in the DS performance of AIM-2D (\textbf{21.64 $\to$ 34.36}) and AIM-BEV (\textbf{28.61 $\to$ 35.86}) when trained on the augmented dataset. Since the RC performance is limited by the expert's performance due to insufficient data support, the improvement is primarily attributed to an increase in IS. This indicates robustness in the AV's performance as it incurs lesser penalties for violating traffic rules or colliding with objects in the scene.

In conclusion, we observe performance improvements for both segmentation and AD due to the context-sensitive augmentations provided by our \ours \space pipeline, which enables the models to generalize better to different visual elements across all subgroups.

\textbf{RQ2: How does the augmentation with synthetic data from \ours\ influence a model's performance across different subgroups?}

Additionally, we evaluate the performance of the segmentation models and AD models across different data subgroups. The results in Tables~\ref{tab:performance_per_subgroup_sem_ad} and \ref{tab:performance_per_subgroup_sem} show that the models fine-tuned on the augmented datasets with the synthetic data samples from \ours\ for segmentation and AD, respectively, consistently outperform those trained on the original datasets alone (or original data with color augmentation for segmentation) across most subgroups. This is observed especially in the case of under-represented subgroups like ``Rain, Dawn / Dusk'' and ``Rain, Night'' for segmentation and ``Rain, Day'' and ``Rain, Night'' for AD. This suggests that the targeted synthetic data obtained from \ours\ for under-represented data subgroups enhances the model's performance on those data subgroups.

\begin{table}
    
    \centering\captionsetup{ singlelinecheck=false}
    \begin{small}
    \begin{tabular}{@{}lcccc@{}}
        \toprule
        \textbf{Subgroup} & \multicolumn{2}{c}{\textbf{DeepDrive}} & \multicolumn{2}{c}{\textbf{Waymo}} \\
        \cmidrule(lr){2-3} \cmidrule(lr){4-5} 
                       & \textbf{CaG} & \textbf{no CaG} & \textbf{CaG} & \textbf{no CaG} \\
                       \midrule
        Clear, Day    & \textbf{162.16} &  202.02 & 152.74 & \textbf{146.99} \\
        Clear, Dawn/Dusk   & \textbf{66.47} &  67.05 & \textbf{150.92} & 160.57 \\
        Clear, Night &  \textbf{211.45} & 273.28 & \textbf{46.77} & 80.84  \\
        Cloudy, Day   & \textbf{134.24} &  148.49 & 118.51 & \textbf{114.93} \\
        Cloudy, Dawn/Dusk  & \textbf{144.94} &  199.48 & \textbf{214.55} & 224.94 \\
        Cloudy, Night & \textbf{152.65} &  246.72 & \textbf{58.12} & 107.57  \\
        Rain, Day   & \textbf{133.96} & 154.72 & 121.69 & \textbf{102.79} \\
        Rain, Dawn/Dusk   & \textbf{199.83} & 229.45 & \textbf{124.35} & 129.68  \\
        Rain, Night & \textbf{291.66} & 349.22 & \textbf{62.21} & 112.75\\
        \bottomrule
    \end{tabular}
    \captionsetup{font=small}
    \caption{\textbf{Comparison of Frechet Distance (FD) with and without caption generation for both datasets (lower is better).} We show that using the caption generation scheme reduces the FD score on CLIP-VIT-L16 features between the generated and the ground truth images. This implies that CaG enhances ControlNet's ability to generate dataset-specific images for each subgroup.}
    \label{tab:cagimpact}
    \end{small}

    \vspace{-1.5em}
\end{table}

\textbf{RQ3: Does the proposed caption generation improve the quality of synthetic images?}

Finally, we evaluate the quality of synthetic images generated by our proposed caption generation scheme using Frechet Distance (FD) \cite{frechet1957distance} as shown in Table ~\ref{tab:cagimpact}. We compute FD scores between the subgroup-specific distributions of the synthetic and ground truth images using the CLIP-VIT-L16 vision encoder model embeddings. The results in Table~\ref{tab:performance_per_npm} indicate that the proposed caption generation scheme using our pipeline lowers the FD scores, which implies that the synthetic images resemble the true data distributions from each test subgroup more closely. We also qualitatively show this result in Figure~\ref{fig:syn_res}. These results justify the benefit of our caption generation scheme to provide captions for fine-tuning and generating images with ControlNet, for segmentation and AD tasks. 

\textbf{Limitations.} This work is limited by a few factors. First, the method is tested on image data with single camera views, which may fail to capture the spatial geometry of multi-view images. Additionally, \ours\ does not generate adversarial synthetic data, which could improve model robustness. Finally, \ours\ assumes that the scenes can be accurately described by a VLM.
\section{Conclusion}

This paper presents \ours, a novel data augmentation pipeline that leverages text-to-image generative models such as ControlNet to address dataset imbalances in AD systems, particularly for under-represented driving conditions. With a novel prompting scheme to provide subgroup-specific semantically dense prompts, \ours \space generates high-quality synthetic data without requiring additional manual labeling, significantly reducing human effort to enhance model performance in under-represented subgroups. We demonstrate the effectiveness of \ours\ through experiments on segmentation and E2E autonomous driving models. Segmentation models fine-tuned on augmented datasets improve by up to 2.3 mIoU, and the performance of AD models is enhanced by 20\% across diverse conditions. While our work is limited to single-view images, in future work, we aim to show how \ours\ can be utilized to improve real-world E2E AD models that generate motion plans from multiple camera views.

{
    \small
    \bibliographystyle{ieeenat_fullname}
    \bibliography{references/references}
}


\clearpage
\setcounter{page}{1}
\maketitlesupplementary

The supplementary section is organized as follows:
\begin{enumerate}
    \item We depict the dataset distribution across all subgroups in all the datasets used in this paper in Section \ref{supp:dataset}.
    \item We elaborate on the performance of Autonomous Driving models (AD) over different subgroups across all AD metrics in Section \ref{supp:perf}.
    \item We provide more qualitative visualizations of the ablation study, with respect to the impact of the subgroup-specific caption generation and the fine-tuning of ControlNet, in Section \ref{supp:abl}.
    \item Finally, we provide qualitative visualizations of the synthetic data samples used for the segmentation and driving tasks in Section \ref{supp:qual}.
\end{enumerate}

\section{Dataset Analysis}
\label{supp:dataset}
For semantic segmentation and AD tasks, the datasets are categorized into subgroups based on weather (\textit{Rain, Clear, Cloudy}) and time of day (\textit{Dawn / Dusk, Morning, Night}). We present the data distribution of the original dataset and the augmented dataset (with synthetic images from \ours) for the  Waymo and DeepDrive datasets, and the AD dataset obtained from the CARLA. We show that the distribution of images across various subgroups on the augmented dataset is more uniform across different subgroups compared to the original dataset.

\begin{figure}[ht]
    \centering   
    \includegraphics[width=0.83\linewidth]{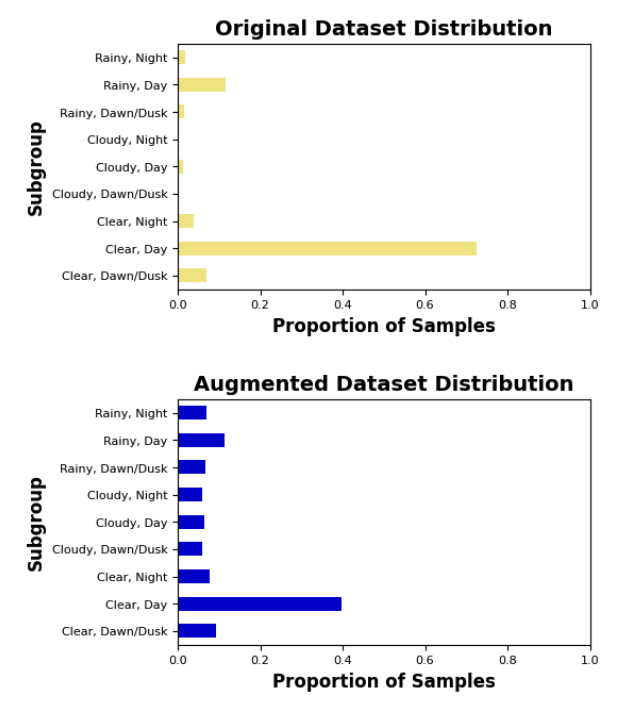}
            \vspace{-0.35cm}
    \caption{The distribution of identified subgroups in the original and augmented dataset for segmentation experiments on the \textbf{Waymo} dataset. }

    \label{fig:dis_fig_waymo}
    \vspace{-0.25cm}
\end{figure}

\begin{figure}[ht]
    \centering   
    \includegraphics[width=0.85\linewidth]{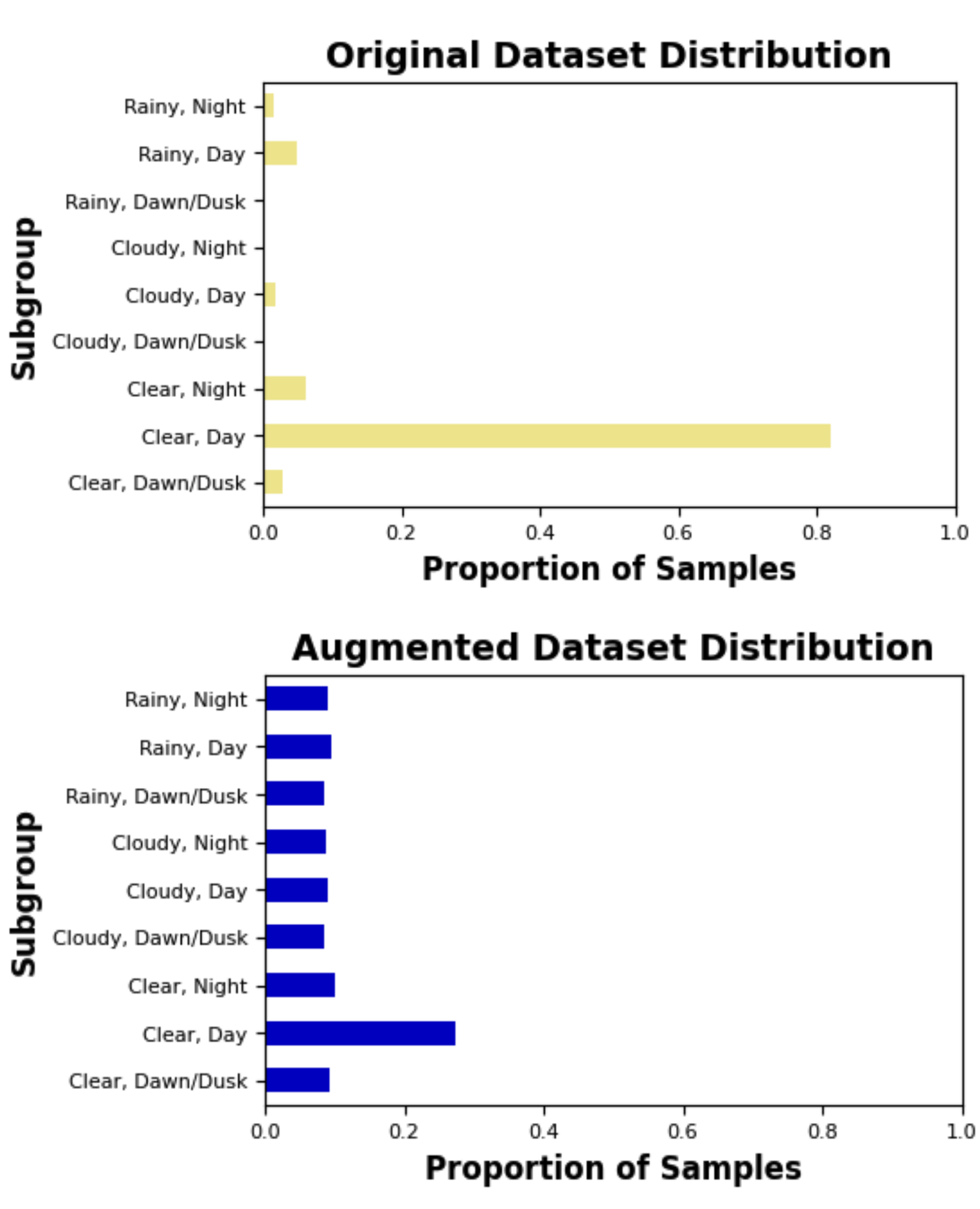}
    \vspace{-0.5cm}
    \caption{The distribution of the identified subgroups in the original and augmented dataset for segmentation experiments on the \textbf{DeepDrive} dataset. }
    \label{fig:dis_fig_bdd}
  \vspace{-0.5cm}
\end{figure}

\begin{table*}[h]
    \centering
    \captionsetup{ singlelinecheck=false}
\label{tab:transposed} 
\caption{\textbf{Performance of AIM-2D across different data-subgroups.}}
\begin{tabular}{@{}lcccccccccc@{}}
\toprule
\multirow{2}{*}{\textbf{Metric}} & \multirow{2}{*}{\textbf{Aug}}&  & \textbf{Clear} &  &  & \textbf{Cloudy} &  & & \textbf{Rain} &  \\
& & \textbf{Twi} &\textbf{Day} & \textbf{Night} &\textbf{Twi} & \textbf{Day} & \textbf{Night} & \textbf{ Twi} & \textbf{Day} & \textbf{Night} \\
\hline 
\textbf{RC} &  & 76.04 & 55.22 &  \textbf{84.91} & 54.05 & 76.05 & 55.02 & 48.65 & \textbf{84.61} & 45.97 \\ 
\textbf{IS} & No & 0.224 & 0.483 & 0.073 &  \textbf{0.729} & 0.204 & 0.727 & 0.259 & 0.244 & 0.352 \\
\textbf{DS}&  & 19.69 & 23.20 & 6.11 & \textbf{37.25} & 18.77 & 43.72 & 14.68 & 23.93 & 3.42\\ 
\midrule
\textbf{RC} &  & \textbf{84.81} & \textbf{55.30} &  84.05 & \textbf{54.98} & \textbf{83.59} & \textbf{55.02} & \textbf{82.58} & 77.02 & \textbf{69.38} \\ 
\textbf{IS}& Yes & \textbf{0.392} & \textbf{0.631} & \textbf{0.312} & 0.339 & \textbf{0.373} & \textbf{0.791} & \textbf{0.266} & \textbf{0.551} & \textbf{0.472}\\ 
\textbf{DS} &  & \textbf{39.04} & \textbf{40.30} & \textbf{29.02} & 19.11 & \textbf{33.44} & \textbf{46.32} & \textbf{16.68} & \textbf{50.94} & \textbf{34.23} \\ 
\bottomrule

\end{tabular}

    \label{tab:performance_per_subgroup_sem_ad2}

\end{table*}

\begin{table*}[h]
    \centering
    \captionsetup{singlelinecheck=false}
\label{tab:transposed}
\caption{\textbf{Performance of AIM-BEV across different data-subgroups.} }
\begin{tabular}{@{}lcccccccccc@{}}
\toprule
\multirow{2}{*}{\textbf{Metric}} & \multirow{2}{*}{\textbf{Aug}}&  & \textbf{Clear} &  &  & \textbf{Cloudy} &  & & \textbf{Rain} &  \\
& & \textbf{Twi} &\textbf{Day} & \textbf{Night} &\textbf{Twi} & \textbf{Day} & \textbf{Night} & \textbf{ Twi} & \textbf{Day} & \textbf{Night} \\
\hline 
\textbf{RC} &  & 83.48 & 55.18 & 76.31 & \textbf{55.18} & \textbf{100.0} & \textbf{55.18} & 64.51 & 83.14 & 69.21 \\ 
\textbf{IS} & No & 0.436 & 0.589 & 0.038 & \textbf{0.573} & 0.447 & \textbf{0.706} & 0.269 & 0.462 & 0.255 \\
\textbf{DS}&  & 39.78 & 31.42 & 2.73 & \textbf{29.88} & \textbf{44.68} & \textbf{43.22} & 18.07 & 43.73 & 3.97\\ 
\midrule
\textbf{RC} &  & \textbf{100.0} & \textbf{55.28} & \textbf{100.0} & 23.66 & 90.86 & 36.99 & \textbf{85.92} & \textbf{100.0}  & \textbf{69.38} \\ 
\textbf{IS}& Yes & \textbf{0.584} & \textbf{0.706} & \textbf{0.254} & 0.438 & \textbf{0.452} & 0.677 & \textbf{0.374} & \textbf{0.536} & \textbf{0.285}\\ 
\textbf{DS} &  & \textbf{58.37} & \textbf{47.94} & \textbf{25.42} & 14.93  & 44.22 & 35.03 & \textbf{27.52} & \textbf{53.67} & \textbf{15.64} \\ 
\bottomrule

\end{tabular}

    \label{tab:performance_per_subgroup_sem_ad3}

\end{table*}

\begin{figure}[H]
    \centering   
    \includegraphics[width=0.80\linewidth]{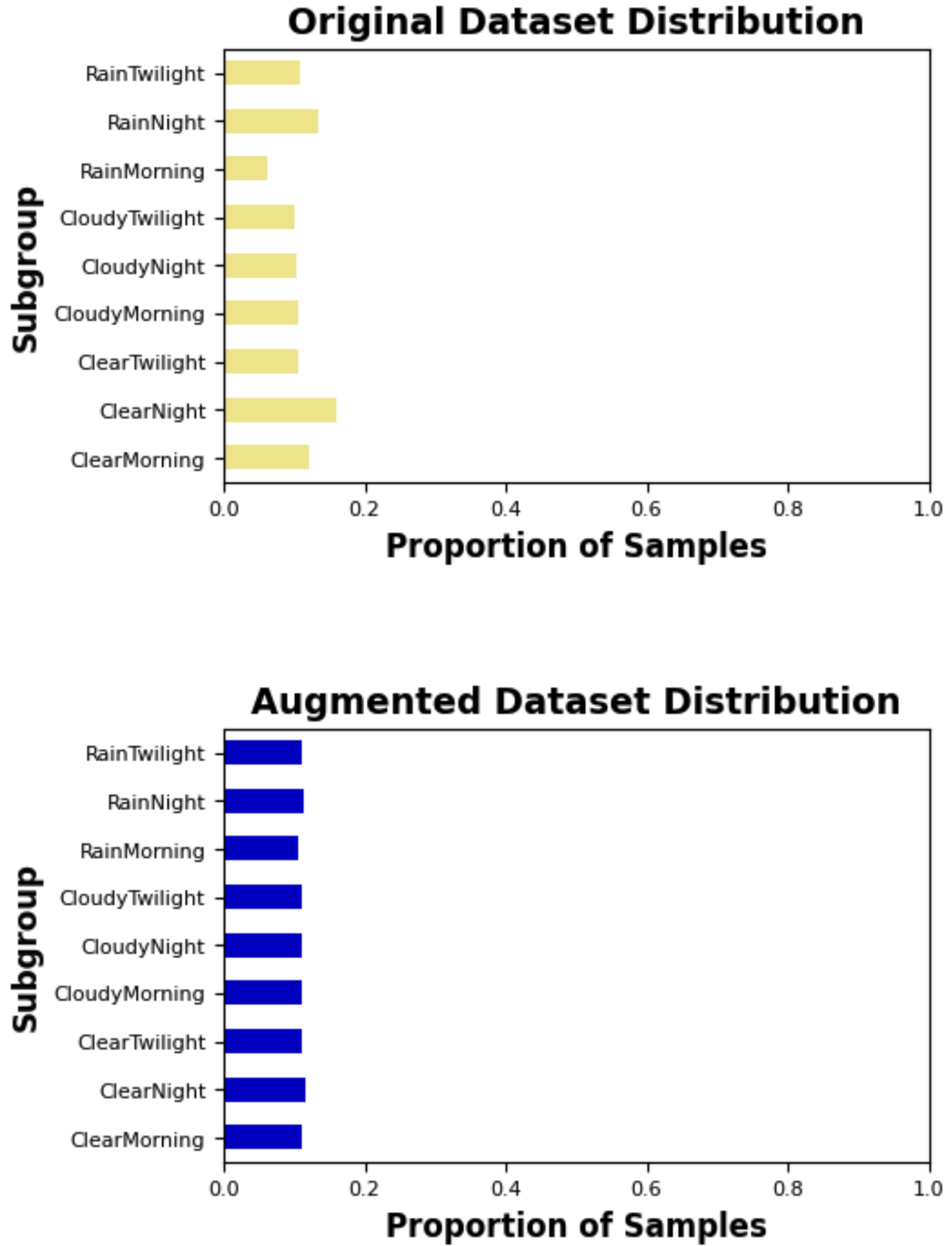}
    \vspace{-0.35cm}
    \caption{The distribution of \textbf{autonomous driving data} collected from CARLA across all identified subgroups. We show the final distribution of data after augmenting the dataset.}
    \label{fig:dis_fig_Carla}
    \vspace{-0.35cm}
\end{figure}


\newpage
\section{Detailed Performance of Autonomous Driving Models}
\label{supp:perf}

We present a detailed breakdown of the Driving Score (\textbf{DS}) as referenced in the main paper. Here, we present the Route Completion (\textbf{RC}) score and the Infraction Score (\textbf{IS}) of the learned AD policies for each data subgroup and model. In the following tables as shown in Table \ref{tab:performance_per_subgroup_sem_ad2} and Table \ref{tab:performance_per_subgroup_sem_ad3}, we report the above metrics for each AD model trained on the original and augmented (with synthetic samples from \ours) datasets. We highlight the best-performing models for each metric across each subgroup. We can see that the synthetic augmentations to the dataset enable autonomous driving models to improve route completion and infraction scores across most subgroups (especially for the AIM-2D model). For AIM-BEV, barring subgroups corresponding to cloudy weather, we see that synthetic data improves the model's performance on all subgroups.

\begin{figure*}
    \centering
    \includegraphics[width=1.0\linewidth]{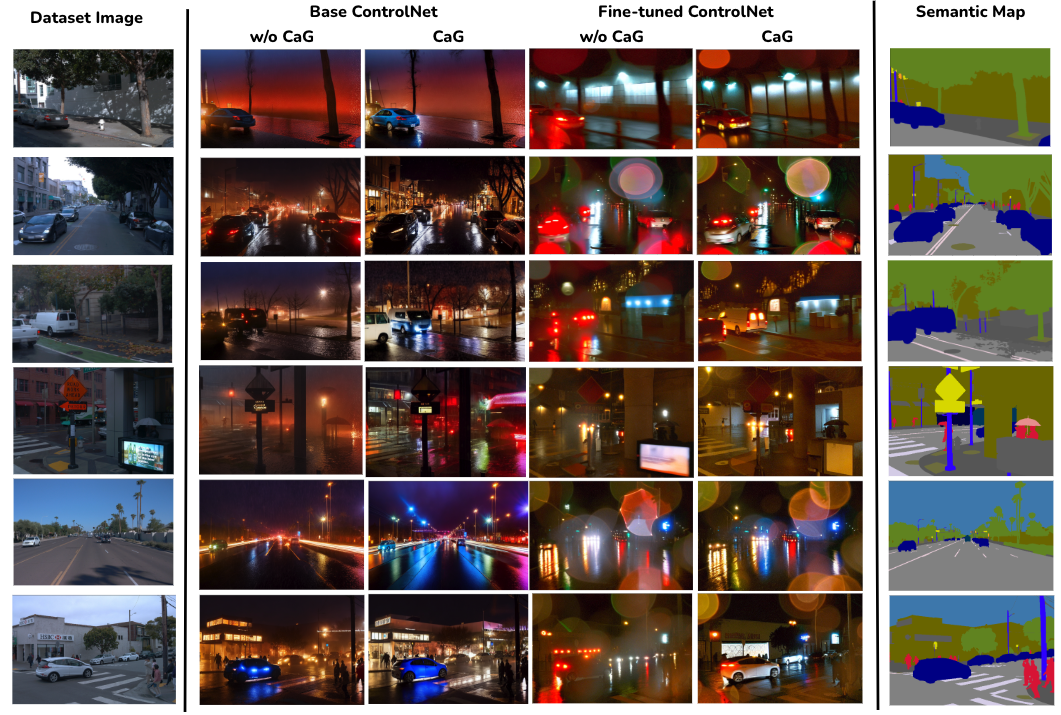}
    \vspace{-0.25cm}
    \caption{\textbf{More qualitative examples for the ablation study.} We generate images using the base ControlNet model and a fine-tuned ControlNet, without and with our subgroup-specific caption generation (CaG).}
    \label{fig:ablation_app}
    \vspace{-0.25cm}
\end{figure*}

\section{Ablation Study}

\label{supp:abl}

We conduct an ablation study to qualitatively visualize the impact of our proposed subgroup-specific caption generation scheme by ControlNet. First, we ablate the impact of fine-tuning ControlNet on the original datasets, on synthetic data generation. Next, we ablate the impact of the effect of caption generation on both variants of ControlNet. We conduct the ablation study for synthetic images generated under ``Rainy and Night'' weather conditions using both ControlNet models. We chose this subgroup specifically as it is the least represented subgroup in the Waymo dataset. The original input images are sourced from the Waymo dataset, captured in ``Clear and Day'' conditions which is the most over-represented subgroup in the dataset. 

We present our key observations as follows:
\begin{enumerate}
    \item
    \textbf{Row 1 (Street Scene):} The base ControlNet model without CaG generates hyper-realistic but generic results, with unnatural backgrounds. Adding CaG improves realism slightly but fails to capture finer details like trees. The fine-tuned model without CaG introduces dataset-specific features, such as raindrop blurring on the car windows, but still exhibits artifacts like distorted vehicle shapes. With CaG, the fine-tuned model produces a more cohesive scene, maintaining structural integrity while accurately simulating rainy conditions.
    \item
    \textbf{Row 2 (Urban Street):} Without CaG, the base ControlNet model generates artificial lighting effects that lack alignment with real-world urban environments. With CaG, the lighting becomes more natural and is well-aligned with the semantic mask. The fine-tuned model without CaG introduces effects due to rain, such as reflections, but suffers from generating key semantics such as lane markers. Incorporating CaG eliminates this issue, producing a realistic scene with lane markers and sharp details such as reflections.
    \item
    \textbf{Row 3 (Parked Vehicles in a Residential Area):} The base ControlNet model without CaG generates images with overly darkened areas. Adding CaG slightly improves lighting consistency but fails to generate vegetation and buildings. The fine-tuned model without CaG captures rainy weather effects more effectively but introduces distortions in vehicle shapes and road markings. With CaG, the scene becomes significantly more realistic, outlining vehicles, vegetation, and buildings clearly.
    \item \textbf{Row 4 (Work Zone):} The base ControlNet model struggles to replicate realistic work zone conditions. The fine-tuned model introduces a more realistic background. CaG further enhances these results by preserving semantic elements like signboards and people with umbrellas, therefore aligning the scene with real-world conditions.
    \item
    \textbf{Row 5 (Highway Scene):} Without CaG, both models fail to maintain lane markers or pavement clarity under rainy conditions. The fine-tuned model with CaG produces a much more realistic highway scene, including reflections of streetlights on wet surfaces.
    \item
    \textbf{Row 6 (Urban Intersection):} The base ControlNet model generates inconsistent lighting and reflections on vehicles. Fine-tuning improves realism by introducing natural lighting. With CaG, the scene includes critical features like road markings, buildings, and vehicles.
\end{enumerate}

\paragraph{Limitation:} While \ours\ produces more dataset-specific images that improve the downstream task models, a major deficiency observed in synthetic images is the preservation of text on signboards, as illustrated in Row 4. These might be crucial for end-to-end autonomous driving.

\section{Qualitative Visualizations}
\label{supp:qual}
We attempt to provide qualitative visualizations of the synthetic images obtained for different tasks and datasets. Here, we sample an image and semantic mask pair and showcase its synthetic variants across different data subgroups obtained with \ours. 

\subsection{Semantic Segmentation Examples}

\begin{figure*}[ht]
    \centering
    \begin{subfigure}[b]{0.78\linewidth}
        \centering
        \includegraphics[width=\linewidth]{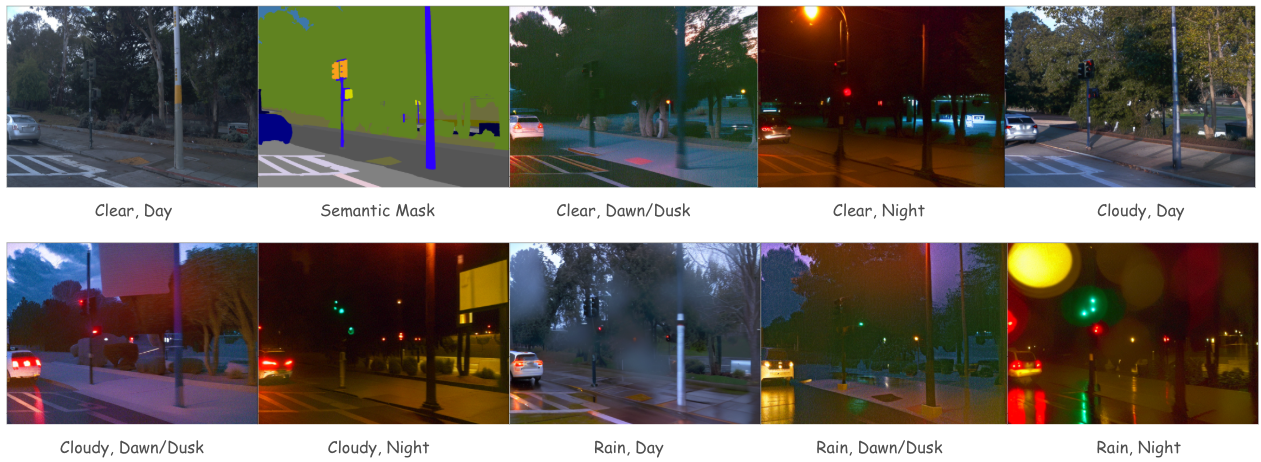}

        \label{fig:syn_vis1}
    \end{subfigure}
    
    \vspace{0.2cm} 
    
    \begin{subfigure}[b]{0.78\linewidth}
        \centering
        \includegraphics[width=\linewidth]{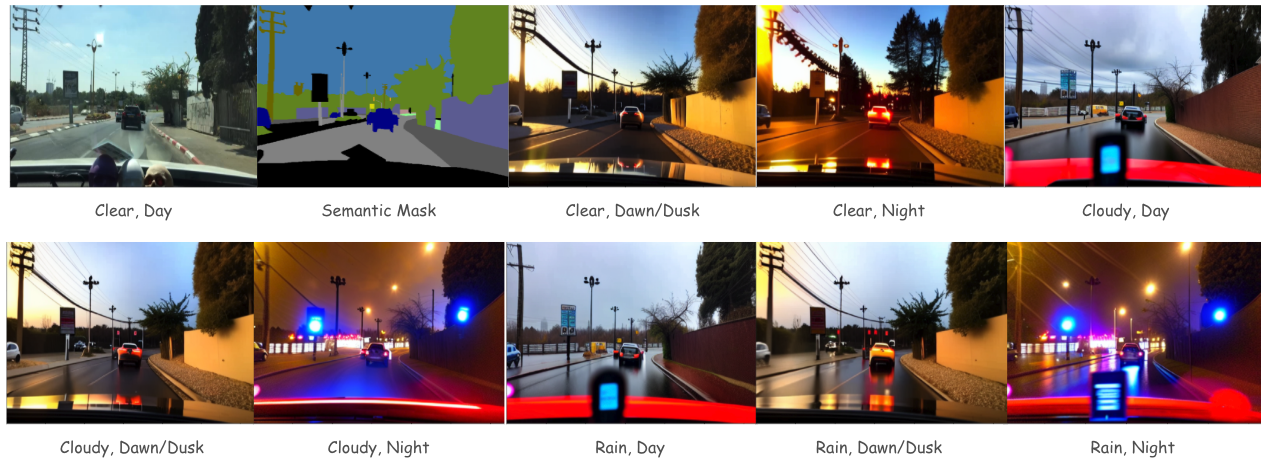}

        \label{fig:syn_vis2}
    \end{subfigure}
    
    \vspace{0.2cm} 
    
    \begin{subfigure}[b]{0.78\linewidth}
        \centering
        \includegraphics[width=\linewidth]{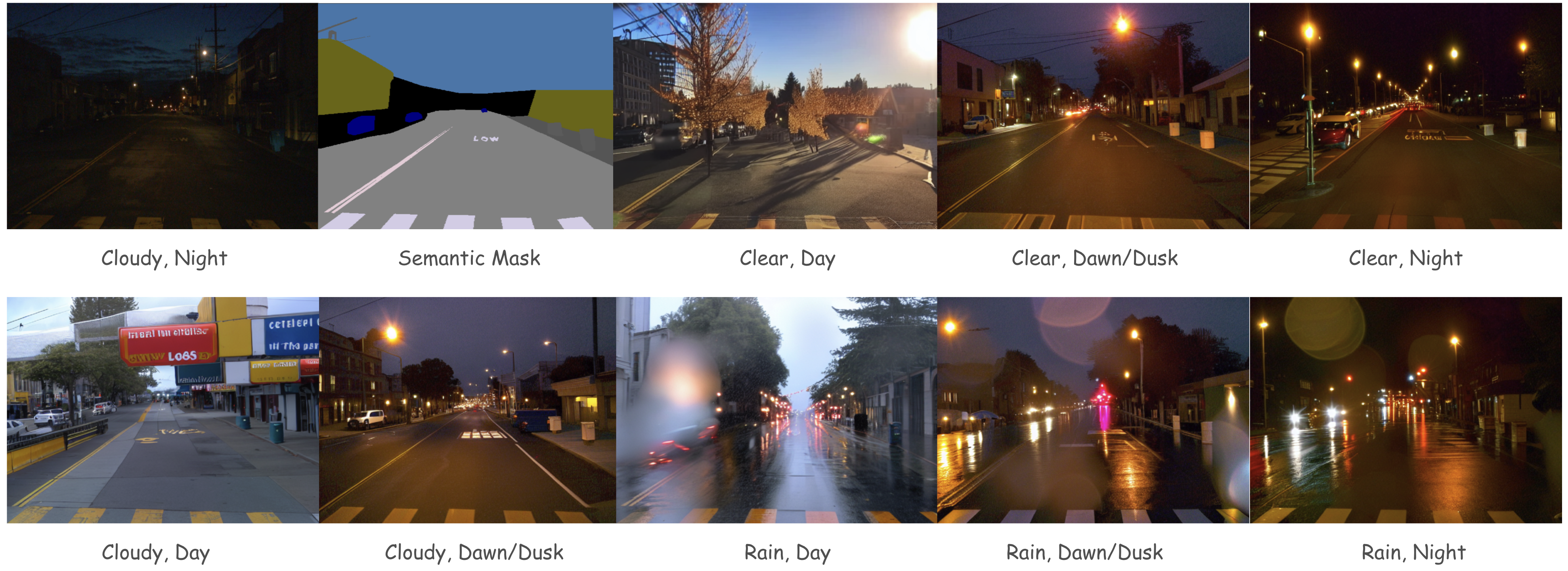}

        \label{fig:syn_vis3}
    \end{subfigure}
    
    \vspace{0.2cm} 
    
    \begin{subfigure}[b]{0.78\linewidth}
        \centering
        \includegraphics[width=\linewidth]{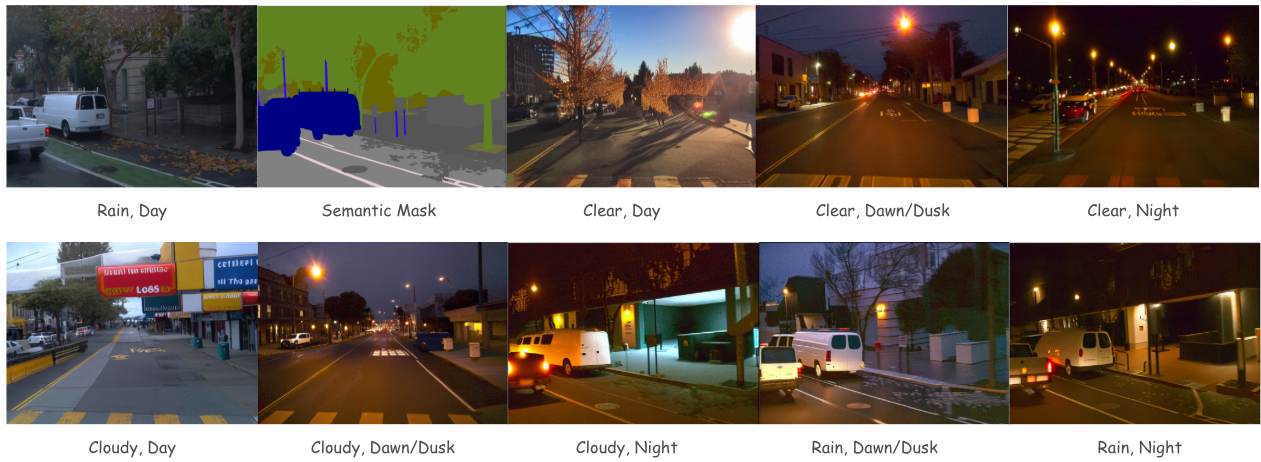}

        \label{fig:syn_vis4}
    \end{subfigure}

    \caption{\textbf{Visualization of synthetic images across different weather conditions and times of day for the semantic segmentation experiments that use the Waymo and DeepDrive datasets.} Every odd row depicts an image and its mask from the dataset followed by its synthetic variants across all conditions.}
    \label{fig:combined_visualization_vertical}
\end{figure*}

We visualize a few more examples across different urban scenes obtained from the Waymo and DeepDrive datasets for semantic segmentation in Figure \ref{fig:combined_visualization_vertical}.

\subsection{Autonomous Driving from CARLA}
\begin{figure*}[ht]
    \centering
    \begin{subfigure}[b]{0.8\linewidth}
        \centering
        \includegraphics[width=\linewidth]{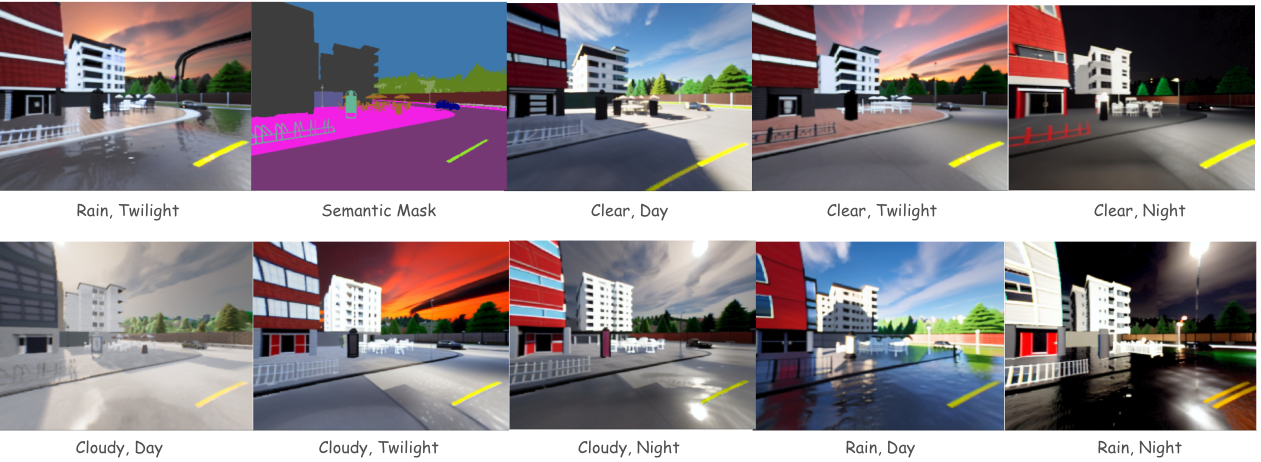}

        \label{fig:syn_carla1}
    \end{subfigure}
    
    \vspace{0.2cm} 
    
    \begin{subfigure}[b]{0.79\linewidth}
        \centering
        \includegraphics[width=\linewidth]{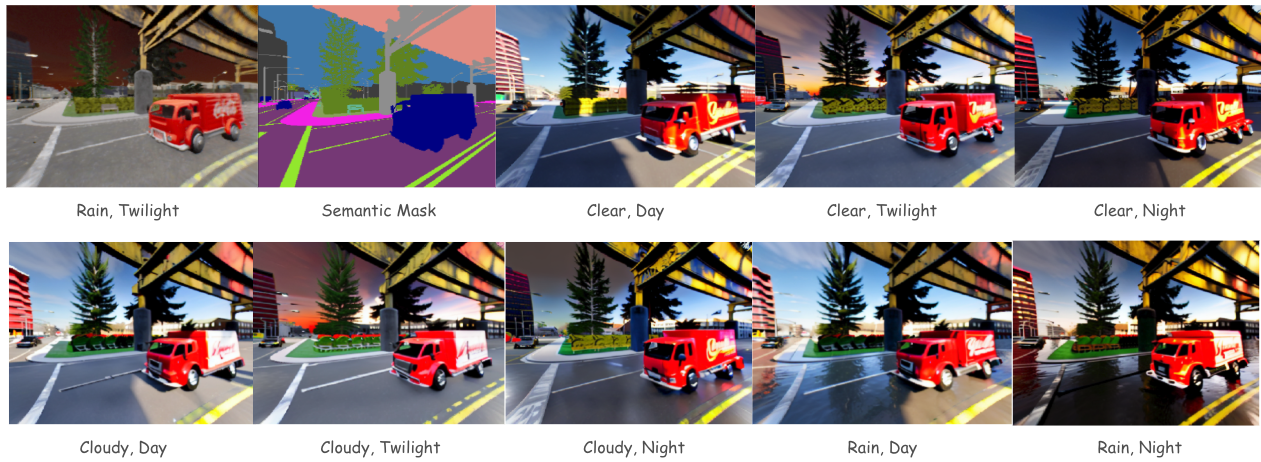}

        \label{fig:syn_carla2}
    \end{subfigure}
    
    \vspace{0.2cm} 
    
    \begin{subfigure}[b]{0.8\linewidth}
        \centering
        \includegraphics[width=\linewidth]{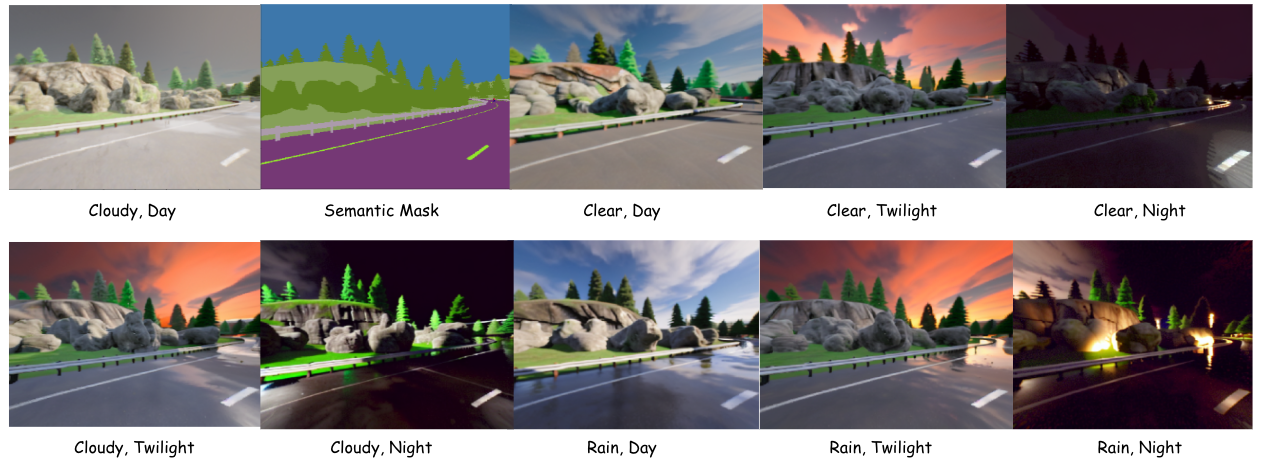}

        \label{fig:syn_carla3}
    \end{subfigure}
    
    \vspace{0.2cm} 
    
    \begin{subfigure}[b]{0.793\linewidth}
        \centering
        \includegraphics[width=\linewidth]{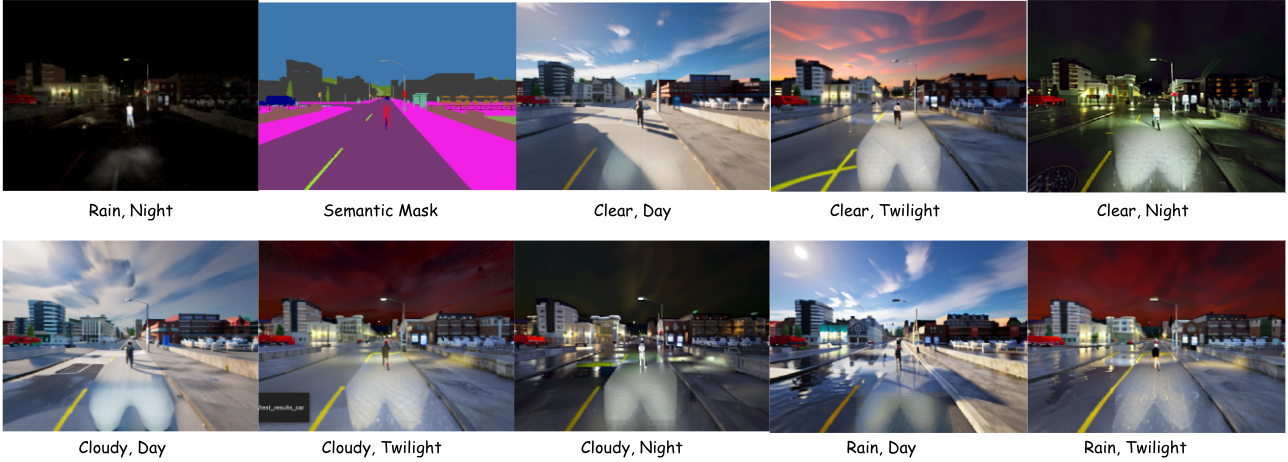}

        \label{fig:syn_carla4}
    \end{subfigure}

    \caption{\textbf{Visualization of synthetic images across different weather conditions and times of day for the end-to-end autonomous driving experiment that uses the CARLA-based dataset.} Every odd row depicts an image and its mask from the dataset followed by its synthetic variants across all subgroups.}
    \label{fig:combined_visualization_vertical2}
\end{figure*}

We also visualize images obtained by \ours\ on the AD dataset obtained from CARLA in Figure \ref{fig:combined_visualization_vertical2}.


\end{document}